\documentclass{article}

\PassOptionsToPackage{numbers,compress}{natbib}

\usepackage[preprint]{neurips_2026}
\usepackage{graphicx}
\usepackage{multirow}
\usepackage{wrapfig}

\usepackage[utf8]{inputenc} 
\usepackage[T1]{fontenc}    
\usepackage{hyperref}       
\usepackage{url}            
\usepackage{booktabs}       
\usepackage{amsfonts}       
\usepackage{nicefrac}       
\usepackage{microtype}      
\usepackage{xcolor}         
\usepackage{siunitx}        
\usepackage{mathtools}      
\usepackage{orcidlink}      

\title{BRICKS: Compositional Neural Markov Kernels for Zero-Shot Radiation-Matter Simulation}

\author{%
  Richard Hildebrandt \orcidlink{0009-0004-4431-0091}\\
  Technical University of Munich\\
  \texttt{richard.hildebrandt@tum.de}\\
  \And
  Evangelos Kourlitis \orcidlink{0000-0001-6568-2047}\\
  Technical University of Munich\\
  \And
  Baran Hashemi \orcidlink{0000-0003-4095-9657}\\
  Max-Planck Institute for\\
  Mathematics in the Sciences, \\
  Leipzig, Germany\\
  \And
  Manuel Bünstorf \orcidlink{0009-0001-4673-9180}\\
  Technical University of Munich\\
  \And
  Thierry Meyer \orcidlink{0009-0008-8099-7213}\\
  Technical University of Munich\\
  \And
  Nikola Boskov \orcidlink{0009-0005-6024-5442}\\
  Technical University of Munich\\
  \And
  Michael Kagan \orcidlink{0000-0002-3386-6869}\\
  SLAC National Accelerator Laboratory\\
  \And
  Dan Rosenbaum \orcidlink{0009-0008-9558-3195}\\
  Haifa University\\
  \And
  Sanmay Ganguly \orcidlink{0000-0003-1285-9261}\\
  Department of Physics\\
  Indian Institute of Technology, Kanpur\\
  \And
  Lukas Heinrich \orcidlink{0000-0002-4048-7584}\\
  Technical University of Munich\\
  Munich Center for Machine Learning\\
  \texttt{l.heinrich@tum.de}\\
}

\begin{document}

\vspace{-1.5cm}
\maketitle

\begin{abstract}
  We introduce a new strategy for compositional neural surrogates for radiation-matter interactions, a key task spanning domains from particle physics through nuclear and space engineering to medical physics. Exploiting the locality and the Markov nature of particle interactions, we create a \emph{next-particle prediction} kernel using hybrid discrete-continuous transformer models based on Riemannian Flow Matching on product manifolds. The model generates variable-sized typed sets of particles and radiation side effects that are the result of the interaction of an incident particle with a material volume. The resulting kernel can be composed to simulate unseen large-scale material distributions in a zero-shot manner. Unlike mechanistic simulators, our model is designed to be differentiable, provides tractable likelihoods for future downstream applications. A significant computational speed-up on GPU compared to CPU-bound mechanistic simulation is observed for single-kernel execution. We evaluate the model at the kernel level and demonstrate predictive stability over multi-round autoregressive rollouts. We additionally release a novel 20M-event radiation-matter interaction dataset for further research.
\end{abstract}

\newpage

\section{Introduction}

\begin{figure}[ht]
      \centering                                                        \vspace{-0.2cm}
      \includegraphics[width=0.83\textwidth]{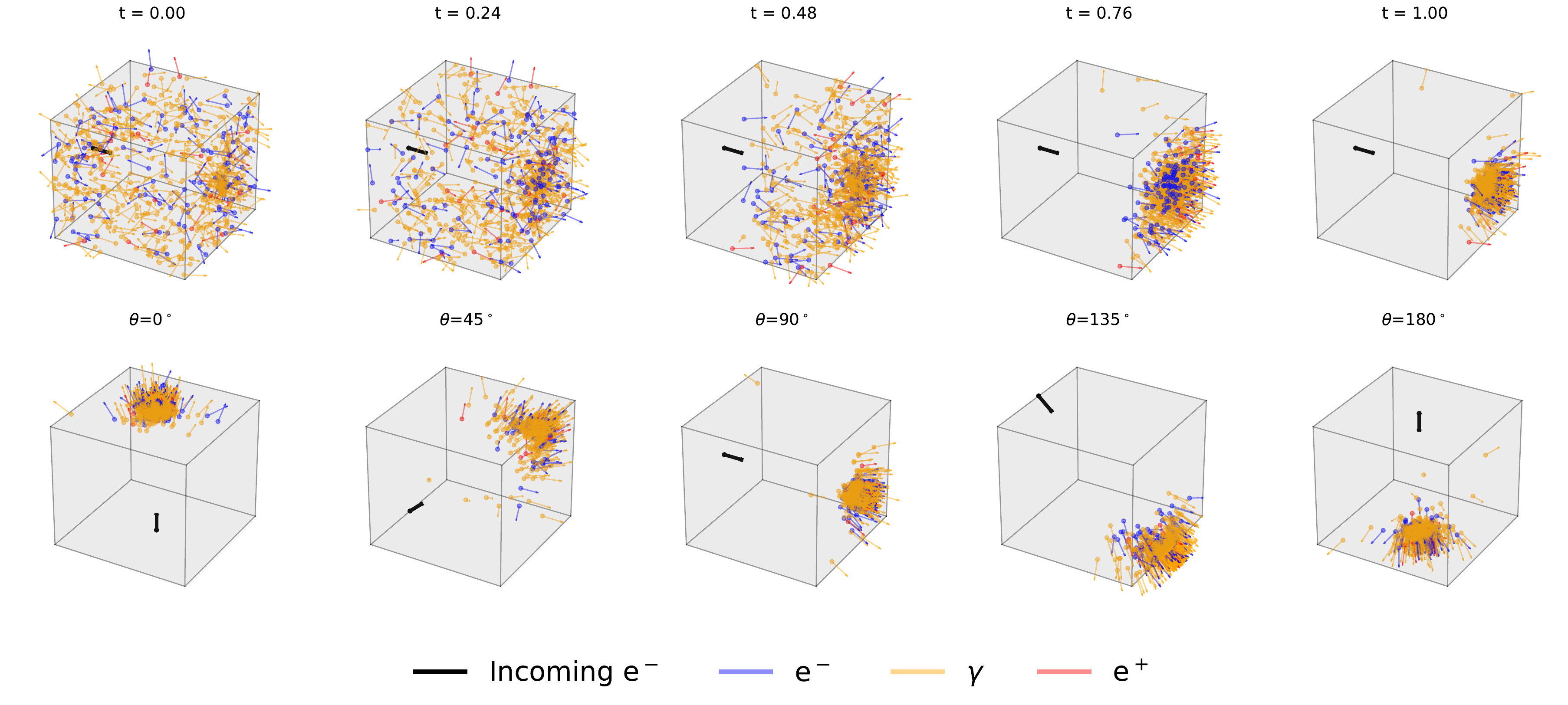}
      \caption{The BRICKS Model: We train a Markov Kernel for Radiation-Matter Interactions. The model describes the effects of particles incident on a volume of material and generates secondary particles as well as material side effects conditional on an incident particle as well as the material properties of the volume. Top: Flow Matching Integration trajectory from $t=0$ to $t=1$. Bottom: Generated Secondaries for varying incoming particle conditions.}
      \label{fig:page1}
\end{figure}   

Composition is broadly seen as a key to generalization: If a model successfully learns a small family of concepts and learns (or is equipped) with a rule on how to compose and arrange those concepts into larger whole, the overall number of instances the model can generate or describe grows exponentially. In language, even a small vocabulary can generate an enormous number of texts when combined with a composition rule like next-word-prediction. In physics, the understanding of how a single atom works and how bonds are formed (i.e. how atoms are composed) leads to the rich field of chemistry. Crucially, if the composition rule and basic building blocks are understood well, it is not necessary to be exposed to composed instances at training time, which leads to zero-shot generalization.

This principle also applies to the problem for simulating the interactions between radiation and matter, a key task in fields as varied as particle physics~\citep{ATLAS:2010arf}, space~\citep{SPENVIS} and nuclear engineering~\citep{GARCIA201773} and medical physics~\citep{Med_MC}. Unlike classical physics, these interactions are random due to their quantum nature: rather than integrating time-steps of differential equations, simulation can be viewed as an iterative stochastic generative task. In fact, this is the approach used extensively in mechanistic simulators~\citep{AGOSTINELLI2003250}, which track particles by sampling random, possibly one-to-many, local interactions from a  `basic vocabulary' of quantum interactions. Crucially, such interaction trees are strictly Markovian: the next transition of a particle only depends on its parent state and its surrounding material.

Mechanistic simulators, however, come with key drawbacks: Closed-form probabilities for stochastic transitions are only available at very short length-scales. This leads to a very high number of iterations and computational cost to complete a simulation; for instance, Large Hadron Collider experiments are projected to use 20-30\% of their CPU budgets to simulate radiation effects from high energy particles over the next decade~\citep{CERN-LHCC-2022-005}. Such simulators are also implemented as pure samplers, whereas it has been shown~\citep{doi:10.1073/pnas.1915980117} that access to the likelihoods and gradients of the iterated stochastic rollouts can significantly improve downstream tasks. 

The advances in generative modelling with neural networks are widely seen as an avenue for creating surrogate models of physics simulation in general~\citep{sanchezgonzalez2020learning} and radiation-matter simulation in particular~\citep{hashemi2024taxonomic}, promising significant speed-up as well as access to likelihoods and gradients. However, currently radiation-matter surrogates are trained on full-scale iterated rollouts of mechanistic simulators and thus lose the compositional nature and zero-shot capabilities of mechanistic simulators in turn.

In this work, we aim to advance towards combining both ideas and distill the radiation-matter interaction physics into a \emph{composable surrogate model} that accelerates simulation, provides access to likelihood and gradient data while maintaining zero-shot generalization. We train a \emph{Next-Particle Markov Kernel} that models the effect of an incident particle with an extended region of material and yields secondary particles emerging again from this region as well as side effects in the material itself. Because the kernel effectively models the collective effect of a large number of microscopic quantum transitions, many fewer iterations are required in a rollout, which may yield significant acceleration~\footnote{This approach may be compared to the difference of e.g. character-level vs. token or word-level language models. Unlike language, however, we can exploit the strict Markovian nature of physics and train on a ``context-window'' of size 1.}. Our choices of neural network architectures are further designed to ensure access to likelihoods and gradients for future work. While a fully featured realization of this vision requires a large-scale effort, we report three concrete novel contributions that showcase the feasibility of this approach towards building general-purpose neural surrogates for particle-matter simulation.

\begin{enumerate}
    \item We develop a composable multi-factor probabilistic model for particle-matter interactions.
    \item We introduce \texttt{CaloBricks}, a novel large-scale dataset of particle-material interactions for multiple material and particle conditions.
    \item We introduce and train a mixed discrete-continuous generative model BRICKS (\textbf{B}roadly \textbf{R}eusable \textbf{I}ntera\textbf{C}tion \textbf{K}ernel \textbf{S}urrogates) using the developed physics model and evaluate it both as a standalone model and as a Markov kernel under autoregressive rollouts.
\end{enumerate}

\section{Probabilistic Particle Matter Interaction Model}

\begin{figure}[h]
      \centering                                                                
      \includegraphics[width=\textwidth]{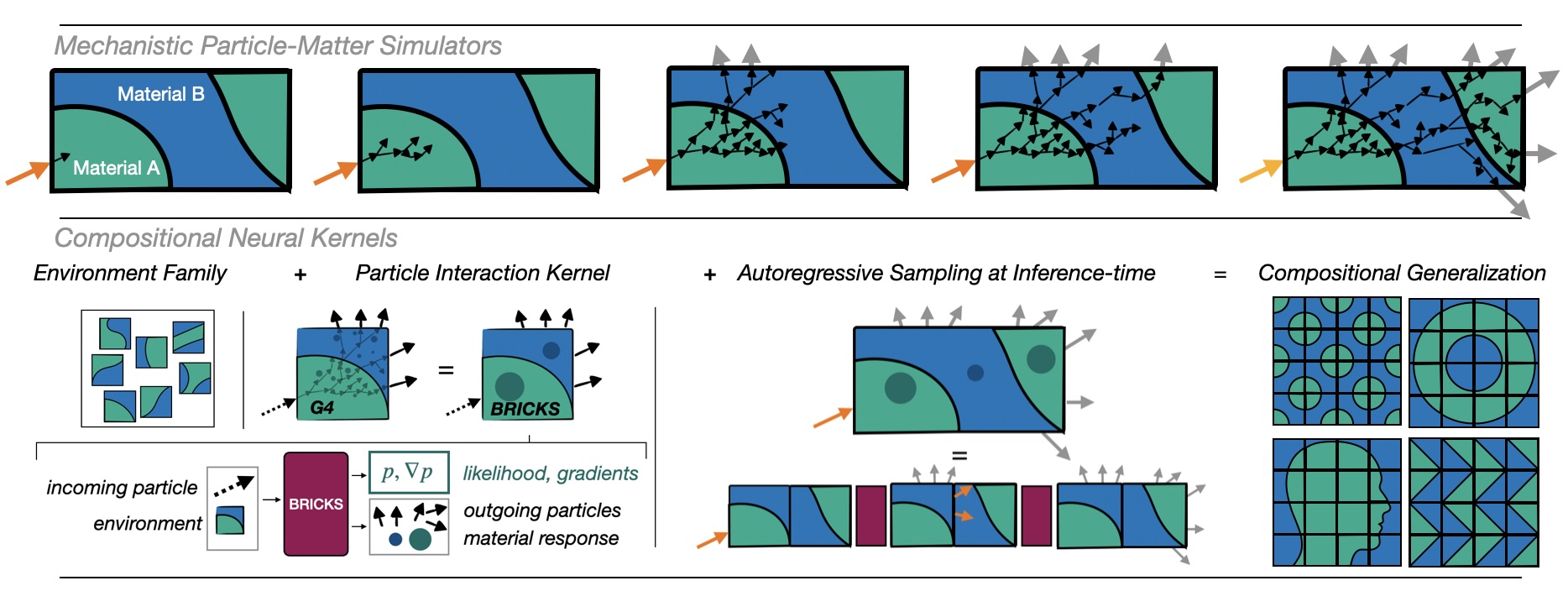}
      \caption{\textbf{Compositional neural kernels for particle transport.}
      \emph{Top:} Mechanistic Monte Carlo Simulators of particle matter interactions iteratively track particles as they undergo Markovian microtransitions.
      \emph{Bottom:} The BRICKS model adopts the same underlying causal structure, but distills the microphysics into medium-scale  neural interaction kernels unlocking both simulation speed-ups as well as access to likelihoods and gradients. Unlike other neural surrogates, the compositional nature of such neural Markov kernels enables zero-shot generalization to unseen material distributions.}
      \label{fig:overview}
\end{figure}

We propose a probabilistic model that is inspired by the simulation model, used in mechanistic simulators, but uses neural components with tractable likelihoods and generative capabilities for every stochastic transition. The overall concept is summarized in Figure~\ref{fig:overview}.

In this model, the large-scale material distribution (or ``geometry'') $G$ is assumed to be composed of elementary (possibly continuously parameterized) family $\Lambda$ of material building blocks. A building block $\lambda \in \Lambda$ may be described by a mixture of discrete and continuous features such as material (discrete) or density (continuous). Even with a small family of building blocks, an enormous number of large-scale geometries can be constructed thanks to the combinatorial nature of composition.

Like mechanistic simulators, we model the simulation of particles as a sequence of probabilistic transitions $s_1\to s_2, \dots s_n$, where each state $s_i$ consists of a set of particles $\{z\}_i$ in an environment $\lambda_i$ and a global material response $E_i$, $s_i = (\{z\}_i, E_i)$. A simulation then consists of executing the transitions until a state with an empty particle list is reached.

To transition from $s_i$ to $s_{i+1}$, each particle $z^k_i$ from the set $\{z\}_i$ undergoes a one-to-many transition $z_i^k \to \{z^{1|k}_{i+1},\dots, z^{n|k}_{i+1}\}$ where the resulting particle set is of variable size and potentially empty. In addition this transition produces a side effect $\delta^k_{i+1}$. The transition probabilities further depend on material properties $\lambda_i$ associated with the environment surrounding the particle $z^k_i$. In total, an incoming particle $z_\mathrm{in}$ thus undergoes a transition described by a kernel of the form $p(\{z\}_\mathrm{out}|z_\mathrm{in},\lambda)$. The particle set of the state $s_{i+1}$ is then constructed from the union of the variable-sized sets produced by each of the prior state's particles. Likewise, the global material response $E_{i+1} \coloneq E_{i+1}(E_i,\{\delta\}_{i+1})$ is constructed by an application-specific accumulation function using the prior state $E_i$ and the set of side effects $\{\delta\}_{i+1}$ produced in the individual transitions of this iteration round. The core of the simulation model is thus the single-particle transition kernel $p(\{z\}_\mathrm{out},\delta|z_\mathrm{in}, \lambda)$. For conciseness in notation, we combine the two conditioning variables into the symbols $\theta = (z_\mathrm{in},\lambda)$ and $y = (\{z\}_\mathrm{out},\delta)$.

We posit that in order to create a surrogate of particle-matter simulation for a large-scale geometry $G$ it is not necessary to ever run a full geometry rollouts of the mechanistic simulator. Instead, it is sufficient to create a high-precision surrogate for the local transition kernel $q_\phi(y|\theta) \approx p(y|\theta)$, where $\phi$ are learnable parameters. To enable autoregression, the model must be amortized over all $\theta$. Once the local kernel is trained, then any geometry that is composed out of the building blocks in $\Lambda$ can be simulated in a zero-shot manner through autoregressive application.

As a first demonstrator of the above model for a particle transition kernel, we model the interactions of particles with a cubic environment made up of a homogeneous material with varying density. As a starting point, we first consider so-called electromagnetic cascades, i.e. applications in which the dominant particles created during the transitions are limited to photons ($\gamma$), electrons ($e^-$) and positrons ($e^+$). In our simple model, the side effect $\delta$ is just the energy deposited into the cube material. In future work, the same model can seamlessly be extended to more complex material building blocks and a broader set of particles and richer side effects such as highly granular internal energy deposition maps. Using these building blocks zero-shot generalization covers all composed voxelized volumes made up of such variable-density cubes.

\section{Related Work}

Deep generative models as surrogates for radiation-matter simulation have been extensively studied in particle physics \citep{hashemi2024taxonomic}. The dominant paradigm has been to learn a surrogate for a \emph{fixed} geometry from full-length rollouts from mechanistic simulators. Seminal examples span GAN-based shower generators such as CaloGAN~\citep{paganini2018calogan}, normalizing-flow surrogates such as CaloFlow~\citep{krause2021caloflow}, and more recent score based and flow-matching based models such as CaloScore v2 and CaloDREAM~\citep{mikuni2024caloscorev2,favaro2025calodream}. The same line of work has also been pushed to more demanding resolutions with ultra-high-granularity detector simulation using explicit relational inductive biases~\citep{hashemi2024ultrahigh}. These results establish that high-fidelity neural detector surrogates are possible, but in most cases the learned model remains tied to a specific global detector geometry or a fixed parameterized family of responses.

A second line of work has started to relax these assumptions. This includes multi-parameter conditioning of shower generators~\citep{diefenbacher2023newangles}, meta-learning across calorimeter geometries~\citep{salamani2023metahep}, latent or learned geometry adaptation~\citep{amram2023calodiffusion,nguyen2026differentiable}, and representation choices that move beyond fixed voxel grids towards point clouds and irregular graphs~\citep{buhmann2023caloclouds,kobylianskii2024calograph,hashemi2024deep,buss2026caloclouds3}. Related efforts have begun to parameterize detector conditions outside the calorimeter volume itself, such as in ParaFlow~\citep{erdmann2025paraflow}. These are important steps towards more reusable fast simulation. However, they still primarily target end-to-end detector response emulation. In other words, the object being learned is usually still the response of a particular detector system, possibly with conditional flexibility, rather than a local physical transition rule that can be recursively composed on unseen material arrangements.

Our work instead targets a different abstraction level, namely a probabilistic \emph{particle-material interaction kernel} that maps one incident particle and a local material descriptor to a variable-size typed set of outgoing particles and with a material side effect. In this sense, it is closer in spirit to learned simulators based on local interaction rules and autoregressive rollouts~\citep{sanchezgonzalez2020learning,brandstetter2022message,koehler2024apebench} than to end-to-end surrogates, while remaining firmly grounded in transport physics. The closest recent point of contact we are aware of outside collider calorimetry is Generative Monte Carlo~\citep{farmer2025gmc}, which also casts transport as a learned conditional generation problem for constant-cost simulation of cell transitions in the linear Boltzmann setting. Our setting differs in several crucial respects. We model full one-to-many particle transitions with mixed discrete-continuous outputs, explicit likelihoods and differentiability, and we use the learned local kernel in composition to obtain zero-shot generalization to unseen large-scale material geometries. To the best of our knowledge, this combination of \emph{locality}, \emph{set-valued generative transport}, and \emph{zero-shot compositional reuse across unseen material distributions} has not been addressed in prior neural surrogates for radiation-matter simulation.

\section{Model and Training} \label{sec:probmodel}

We implement the particle-matter interaction kernel as a hybrid discrete-continuous multi-modal conditional generative model with tractable likelihoods. The conditioning input modalities are i) the material properties $\lambda$ described by its density $\rho$ and ii) the incoming particle $z$. The particle is identified through its particle type, its position on the cube, its momentum direction and its energy. The output modalities are i) a scalar material response (the energy deposited into the environment) $\delta$ and ii) a variable-size unordered set of outgoing particles $\{z\}_\mathrm{out}$, which are described by a) position, b) momentum and c) the particle type. That is, both incoming and outgoing particles are described by a mixture of discrete and continuous features. In particular, the outgoing particles $\{z\}_\mathrm{out}$ constitute a \emph{set of sets}: $\{z\}_\mathrm{out} \to \{\{z\}^a_\mathrm{out}, \{z\}^b_\mathrm{out}, \{z\}^c_\mathrm{out}\dots\}$, where each subset describes particles of the same type. The resulting density model is equivariant with respect to permutations of output particles of the same type but not of particles of different types.

To account for the structural difference between discrete and continuous features, we implement the particle-matter interaction kernel using two distinct transformer-based neural networks, the first for the discrete features, and the second for the continuous features. Denoting the particle feature set $z = (x,w)$, where $x$ are the continuous features and $w$ is the discrete particle type, the model reads $p(\delta, \{x\}, \{n\} | \theta)  = p(\{n\}|\theta)\,p(\delta, \{x\}|\{n\},\theta)$, where the first factor is a conditional probability density over the per-type sub-set cardinalities, $\{n_a, n_b,\dots n_K\}$, where $K$ is the number of particle types considered, and the second factor is a model for the continuous energy deposition and particle features given the overall condition $\theta$ and a set of sub-set cardinalities. At inference time the ancestral sampling of both factors yields the desired generative model.

\subsection{Cardinality Model}

For the cardinality model $q_\phi(\{n\}|\theta) = q_\phi(n_a,n_b,\dots n_K|\theta)$, we choose a causal autoregressive transformer model. Instead of predicting a variable-size sequence of $n$ particle types (for example: $e^+e^-\,, e^-e^+,\gamma\gamma$ ), we exploit the permutation invariance between same-species particles and predict the particle-type cardinalities directly in a specified order $[n_{e^-},n_{e^+},n_\gamma]$. The resulting cardinality sequence is therefore fixed-length and much shorter than the per-particle sequence, which alleviates the inference-time cost of autoregression. 

The input condition is encoded using two conditioning tokens that act as a prompt for the sequence to be generated. The first token represents the discrete particle type condition using a simple embedding layer and the second token is used for the continuous condition features. The continuous features originate from a manifold $\mathbb{R} \times \mathcal{S}^2 \times \mathbb{R}^3$ (material density, incident position\footnote{here, the incoming position on a cube surface is homeomorphic to a position on a surrounding sphere, hence $\mathcal{S}^2$}, and particle momentum respectively) and are embedded after preprocessing using a linear layer into a continuous vector representation. In addition to the prompt token, the particle type is additionally added as conditioning variable in the transformer per-block following the AdaLN~\citep{xu2019understandingimprovinglayernormalization} method.

The model is trained using teacher forcing and cross-entropy loss, where the categories represent the three sub-set cardinalities. For training, the Schedule-Free variant of AdamW~\citep{defazio2024road} is used with  $\beta = (0.95,0.999)$, learning rate $10^{-3}$ and weight decay $10^{-2}$. At inference time, the sub-set cardinalities are sampled autoregressively.

\subsection{Continuous Feature Model}

For the continuous feature model $q_\phi(\delta,\{x\}|\{n\},\theta)$ we choose as a Riemannian Conditional Flow Matching~\citep{chen2024flowmatchinggeneralgeometries} architecture using a transformer~\citep{vaswani2023attentionneed} backbone. For the material response $\delta$ and each of the outgoing particles (as determined by the cardinality model at inference time) a token is allocated. The two modality types, material response and particle features, are disambiguated using a modality-specific modulation at the input encoding level.

As a flow matching model, the density $q_\phi(\delta,\{x\}|\{n\},\theta)$ results from an estimation of an instantaneous velocity field $v_\phi(\delta,\{x\},t|\theta)$, which then is integrated using an ODE solver at inference time from $t=0$ to $t=1$ starting from samples of a base distribution $p_0 = p_\mathrm{base}$ to yield samples at the target distribution $q_1\approx p$. In particular, this choice of architecture not only allows the ability to draw samples from the learned distribution but also to evaluate the likelihood $q_\phi(\delta,\{x\}|\{n\},\theta)$ and compute derivatives $\nabla q_\phi(\delta,\{x\}|\{n\},\theta)$, achieving the key desiderata of a probabilistic and differentiable surrogate. We leave a detailed study of those advanced outputs for future work.

Among the possible CFM model families, we choose the Riemannian CFM architecture since the generated random variables live on a non-trivial product manifold $\mathcal{M}_\delta \times (\mathcal{M}_p)^n$. The material response $\delta$ is an element of a finite interval $[\delta_\mathrm{min}, \delta_\mathrm{max}] \subset \mathbb{R}$. Each copy of the particle manifold $\mathcal{M}_p = \mathcal{S}^2 \times \mathbb{R}^3$ is itself a product manifold consisting of the particle position on the cube, or equivalently on a surrounding sphere, ${x_p}\in \mathcal{S}^2$, and particle momentum $x_\mathrm{mom}\in \mathbb{R}^3$. In addition, the transformer backbone provides the permutation invariance which is then only partially broken through explicit token embeddings to achieve the sub-set permutation symmetry. 

The token embeddings are designed to reflect the role of each token using learned embeddings. Each token is embedded using the form $v = Wx + b$. Unlike standard transformers, where a single learnable embedding layer is used, we allocate separate learnable parameter spaces for the weight matrix $W$ and the bias vector $b$ in a role-dependent way. Each token role is defined by a triple $(\mathrm{type}, \mathrm{pdg}, \mathrm{mask})$. First, tokens are differentiated as one of three token types: i) a conditioning token, ii) a material response token and iii) an outgoing particle token. Second, based on the generated partitioning of the outgoing particles into per-species sub-sets, a corresponding number of tokens is equipped with an embedding of the corresponding particle type~\footnote{We use the standard Particle Data Group (PDG) identifiers to identify particles} to differentiate it from particle tokens of another type. Third, tokens are either masked or unmasked, as during training, samples with a varying number of outgoing particles are combined in a single batch and padded to a maximum sequence length. For each token, we therefore associate three \emph{biasing vectors} which are added and jointly embedded into a bias vector representation $b = (b_\mathrm{mask} + b_\mathrm{pdg} + b_\mathrm{type})/3$. Similarly, the embedding matrix is constructed as $W = (W_\mathrm{mask} + W_\mathrm{pdg} + W_\mathrm{type})/3$. Notably, due to the partial permutation invariance no standard positional encoding is used. 

To cast the transformer backbone into a flow matching model, the time conditioning is added in two locations:  First, the embedded token vector is augmented additively with a sinusoidal time embedding $\tau(t)$ with alternating $\sin(\omega_i t)$ and $\cos(\omega_i t)$ features. Second, the time conditioning is additionally injected into each transformer block according to the AdaLN procedure. Once embedded, the tokens are processed through an encoder-only transformer backbone. For the token positions that represent non-conditioning tokens, the output token of the transformer is projected into the output velocities space to finalize the model.

The model is trained using the conditional flow matching procedure~\citep{lipman2023flowmatchinggenerativemodeling}. That is, a pair of data points $(y_0,y_1)\sim \pi(y_0,y_1)$ is sampled from a \emph{coupling distribution} and a conditional path that interpolates between $y_0$ and $y_1$ is constructed. The coupling distribution is based on the target distribution $p_\mathrm{target}$ and a base distribution $p_\mathrm{base}$, and in its simplest form is the product of the two $\pi(y_0,y_1)  \sim p_\mathrm{base}\cdot p_\mathrm{target}$. The base distribution is usually a simple noise distribution but can be chosen to be any distribution with a tractable probability density. The impact of non-trivial base and coupling distributions are discussed in the appendix. Conditioned on a pair $(y_0,y_1)$, a conditional path and its associated velocity $v_\mathrm{cond}(y,t|y_0,y_1)$ at a randomly chosen time $t$ acts as a regression target for the neural network output with loss $\mathcal{L} = \mathbb{E}_{(y_0,y_1)\sim \pi, t\sim \pi(t)}[(v_\phi - v_\mathrm{cond})^2]$. The model is trained with Schedule-Free AdamW with $\beta = (0.95,0.999)$, learning rate $5 \cdot 10^{-4}$ and weight decay $10^{-2}$. During inference, the choices of ODE solver and step-size significantly impact inference time and sample quality, which are explored in Section~\ref{ssec:timing}.

\section{Datasets}

Standard Datasets in Particle Interaction datasets ~\citep{calochallenge} typically only contain the overall energy depositions into the material for a large-scale geometry. In order to train a composable particle-interaction kernel a new type of dataset is required that also contains outgoing particles that emerge from the simulation of a material volume. 

In this work we therefore introduce the \texttt{CaloBricks} dataset consisting of 20M simulations of particles interacting with a cube of 10 cm edge length made of Argon (Ar), a material frequently used in particle physics applications. As incident particles electrons, positrons and photons are randomly chosen and placed at random positions on the cube surface. A random inward-facing momentum direction is sampled. The energies are sampled in a range of 20-300 MeV and the material density is sampled in a range of \SI{0.5}{\g\per\cm\cubed} to \SI{10}{\g\per\cm\cubed}. The resulting prior (which we refer to as ``random particle gun'' $\pi_\mathrm{gun} = \pi(\theta)$) provides broad coverage of the targeted conditioning space. The simulations are carried out with the Geant4~\citep{geant4} simulation toolkit.

For each simulated interaction, we record the simulation conditions $\theta$, as well as the outgoing particle data and energy deposition into the cube $(\{z\}_\mathrm{out}, \delta)$. The dataset will be made publicly available on Huggingface and will be continuously expanded to more material types such as copper (Cu), lead (Pb), or lead tungstate ($\mathrm{PbWO}_4$), which are commonly used in calorimeter applications.

\section{Evaluation and Results}

The combined model of cardinality prediction and conditional flow-matching yields a high-precision generative model for particle-matter interactions. In Figure~\ref{fig:trained_model_overview} we compare the discrete cardinality (left) and continuous particle and side effect features generated by the model with the ground truth simulator. In general, good agreement is observed for marginal feature distributions as well as two-feature correlations. To assess the performance of our model more quantitatively, we evaluate it at the single-step level as well as its zero-shot performance over multi-step autoregressive rollouts. To our knowledge no published neural surrogate models radiation-matter at the kernel granularity (rather than full-detector response); we therefore anchor our metrics against the within-Geant4 sampling floor  
  (G4-G4) and the untrained base distributions ($p_\text{phys}$, $p_\text{iso}$) used during training.

\subsection{Kernel Performance}

As a baseline set of metrics, we use two-sample tests that quantify the similarity between samples from the ground truth and learned $x_p\sim p(x|\theta_0), x_q\sim q_\phi(x|\theta_0)$ at fixed conditions $\theta_0$. As two-sample test statistics we use the Maximum Mean Discrepancy (MMD)~\citep{gretton2012kernel} and Energy Distance (ED)~\citep{szekely2013energy}. Since the generated data is multi-modal and variable-sized $y = (\{z\},\delta)$ we first form a vector-valued summary statistic that captures the salient features of the generated set. The full summary vector is described in the appendix but features include e.g. first and second order moments of particle positions and momenta as well as the event cardinalities.

Our summary vector may not fully capture distribution shifts. Therefore, we extend the evaluation by a trained neural-network based classifier, which can exploit high-dimensional correlations between two distributions $p(x), q(x)$ and is thus sensitive to subtler distribution shifts. As the densities in question are parametrized, we also parametrize the classifier $f(x,\theta)$. This allows us to use a single trained network for all parameter points $\theta$. A key metric is the area under the curve (AUC): the closer the network is to chance guessing $\mathrm{AUC}=0.5$, the smaller the distributional distance.

\begin{figure}
    \centering
    \includegraphics[width=\linewidth]{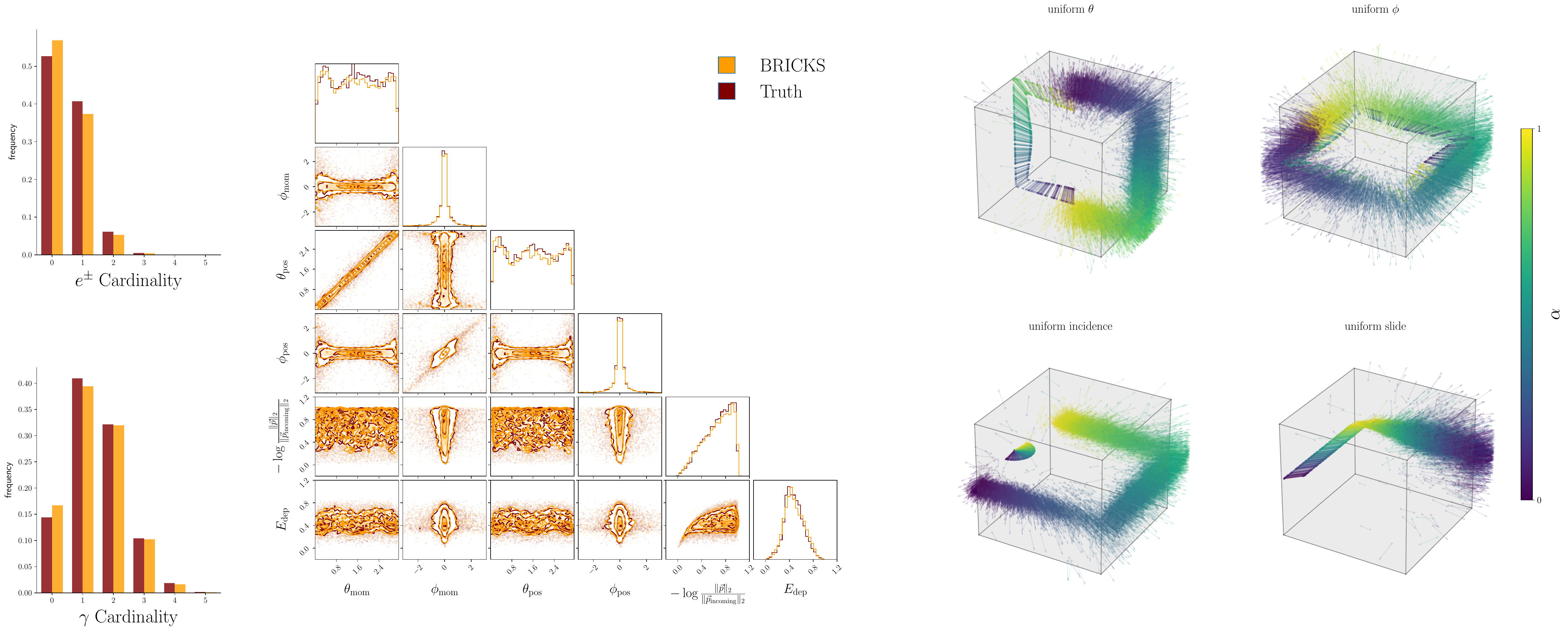}
    \caption{Left: discrete cardinalities and continuous particle and side effect features generated by the BRICKS model compared to the ground-truth mechanistic simulator (G4). Right: visualization of four one-dimensional conditioning-space sub-manifolds.}
    \label{fig:trained_model_overview}
\end{figure}

For zero-shot generalization performance under auto-regression, it is of crucial importance that the model matches the ground truth density well for all possible conditioning vectors $\theta$. We therefore evaluate the metrics for multiple key one-dimensional sub-manifolds in the condition space using uniform prior densities $\pi(\theta)$ on this manifold in addition to the data marginalized over the training data particle-gun prior $\pi_\mathrm{gun}$. In Figure~\ref{fig:trained_model_overview}, four such manifolds are shown: two sweeping over the azimuthal ($\pi_\phi$) and zenith ($\pi_\theta$) angle, one keeping the particle position fixed but varying its incidence angle ($\pi_\mathrm{inc}$) and one keeping the incidence angle fixed but with the incident position $\pi_\mathrm{slide}$ sliding across a face. Two more priors $\pi_E$ and $\pi_\rho$ sample the energy and density respectively.

In Table~\ref{tab:prior_fm}, we report the sample-based and classifier based metrics. To provide a reference scale we compare the achieved values to those between the target distribution and the naive and physical base distribution used in the CFM training. In the appendix, we take an early look at exploiting the tractable likelihood as an additional metric but leave a detailed evaluation for future work.

\begin{table}[h]
  \centering
  \caption{Per-prior quality of the \texttt{BRICKS-M} emulator vs Geant4 (electron beam, $E = 150$\,MeV, $\rho = 3$, pdgid = 11). Point estimate computed on $N_{\text{eval}} = 8{,}000$ events; reported uncertainty is the standard error of the mean (std$/\sqrt{K}$) across $K{=}10$ disjoint sub-samples drawn from the same pool. AUC uses the trained neural-network classifier.}
  \label{tab:prior_fm}
  \vspace{2pt}
  \small
  \setlength{\tabcolsep}{6pt}
  \renewcommand{\arraystretch}{1.15}
  \begin{tabular}{@{}l r r r @{\hspace{6pt}\vrule\hspace{6pt}} r r@{}}
    \toprule
    \textsc{Prior} & \textsc{MMD} & \textsc{ED} & \textsc{AUC} & \textsc{MMD$_{\mathrm{phys}}$} & \textsc{MMD$_{\mathrm{iso}}$} \\
    \midrule
    $\pi_\mathrm{gun}$ & $0.0229 \pm 0.0043$ & $0.0050 \pm 0.0013$ & $0.581 \pm 0.004$ & $0.2087 \pm 0.0024$ & $0.2545 \pm 0.0024$ \\
    $\pi_\varphi$ & $0.0266 \pm 0.0044$ & $0.0061 \pm 0.0012$ & $0.533 \pm 0.006$ & $0.0743 \pm 0.0034$ & $0.2292 \pm 0.0031$ \\
    $\pi_\theta$ & $0.0300 \pm 0.0048$ & $0.0074 \pm 0.0018$ & $0.533 \pm 0.004$ & $0.0780 \pm 0.0034$ & $0.2730 \pm 0.0034$ \\
    $\pi_E$ & $0.0230 \pm 0.0051$ & $0.0052 \pm 0.0017$ & $0.556 \pm 0.004$ & $0.1313 \pm 0.0021$ & $0.3376 \pm 0.0033$ \\
    $\pi_\mathrm{inc}$ & $0.0245 \pm 0.0048$ & $0.0055 \pm 0.0016$ & $0.568 \pm 0.004$ & $0.1710 \pm 0.0024$ & $0.2969 \pm 0.0019$ \\
    $\pi_\mathrm{slide}$ & $0.0230 \pm 0.0050$ & $0.0053 \pm 0.0019$ & $0.575 \pm 0.005$ & $0.1809 \pm 0.0021$ & $0.3637 \pm 0.0020$ \\
    $\pi_\rho$ & $0.0121 \pm 0.0047$ & $0.0014 \pm 0.0016$ & $0.527 \pm 0.004$ & $0.1316 \pm 0.0021$ & $0.2800 \pm 0.0020$ \\
    \bottomrule
  \end{tabular}
\end{table}

\subsection{Inference Time}
\label{ssec:timing}
\vspace{-0.1cm}
\begin{table}[h]
  \centering
  \caption{Inference time and kernel-fidelity (MMD) for BRICKS-M and BRICKS-S at $\rho=10\,\mathrm{g/cm}^3$ on a single H200 GPU, with Geant4 CPU reference timings.}
  \label{tab:pareto_combined_rho10}
  \vspace{2pt}
  \small
  \setlength{\tabcolsep}{5pt}
  \renewcommand{\arraystretch}{1.15}
  \begin{tabular}{@{}ll cc cc@{}}
    \toprule
     & & \multicolumn{2}{c}{\textsc{BRICKS-M}} & \multicolumn{2}{c}{\textsc{BRICKS-S}} \\
    \cmidrule(lr){3-4} \cmidrule(lr){5-6}
    \textsc{Method} & \textsc{Step} & ms/evt & MMD & ms/evt & MMD \\
    \midrule
    \multicolumn{6}{@{}l}{\textsc{Geant4} (CPU): $0.06$ ($\rho{=}1$), $0.35$ ($\rho{=}5$), $0.607$ ($\rho{=}10$) ms/evt} \\
    \midrule
    \multirow{2}{*}{Euler}
      & 2 & $0.073$ & $0.146 \pm 0.003$ & $0.038$ & $0.140 \pm 0.002$ \\
      & 4 & $0.135$ & $0.086 \pm 0.003$ & $0.064$ & $0.091 \pm 0.002$ \\
    \cmidrule(lr){1-6}
    \multirow{2}{*}{Midpoint}
      & 2 & $0.135$ & $0.035 \pm 0.005$ & $0.064$ & $0.047 \pm 0.004$ \\
      & 4 & $0.259$ & $0.028 \pm 0.002$ & $0.117$ & $0.039 \pm 0.003$ \\
    \cmidrule(lr){1-6}
    \multirow{2}{*}{RK4}
      & 2 & $0.259$ & $0.028 \pm 0.003$ & $0.117$ & $0.043 \pm 0.002$ \\
      & 4 & $0.506$ & $0.029 \pm 0.004$ & $0.222$ & $0.044 \pm 0.005$ \\
    \bottomrule
  \end{tabular}
\end{table}

\begin{wrapfigure}{r}{0.4\textwidth}
    \centering
  \includegraphics[width=\linewidth, trim=3pt 2pt 4pt 0pt, clip]{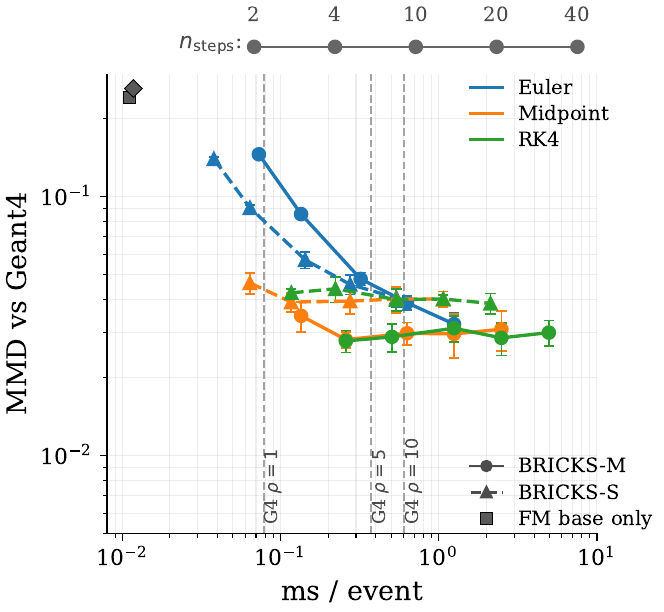}
    \caption{The Timing-Quality Pareto Frontier of the BRICKS model}
    \vspace{-0.5cm}
    \label{fig:timing}
\end{wrapfigure}

A key motivation for neural surrogates of particle simulators is the possibility of significantly speeding up the simulation process. For our model, the inference speed is dominated by the flow-matching component, while the autoregressively generated cardinalities are negligible due to their short sequence length. Beyond the GPU hardware, the  CFM model performance is defined by three choices: i) the overall size of the model ii) the choice of ODE solver and iii) the number of integration steps.
All three potentially trade off accuracy against speed. We therefore characterize the Pareto frontier of the CFM model by comparing multiple combinations in Figure~\ref{fig:timing}.
We compare two models BRICKS-M with ca. 30M parameters and BRICKS-S with just below 10M parameters. For ODE solvers we compare three methods: Euler, Midpoint and RK4. we vary the number of integration steps from 2 to 40. Compared to the CPU runtime of the ground-truth simulator, our model achieves significantly shorter wall-clock runtime on GPU hardware, a gap that widens as the material density (and thus the number of microscopic transitions necessary in mechanistic simulation) increases.

\subsection{Composition Performance}

\begin{figure}[h]
\includegraphics[width=\textwidth]{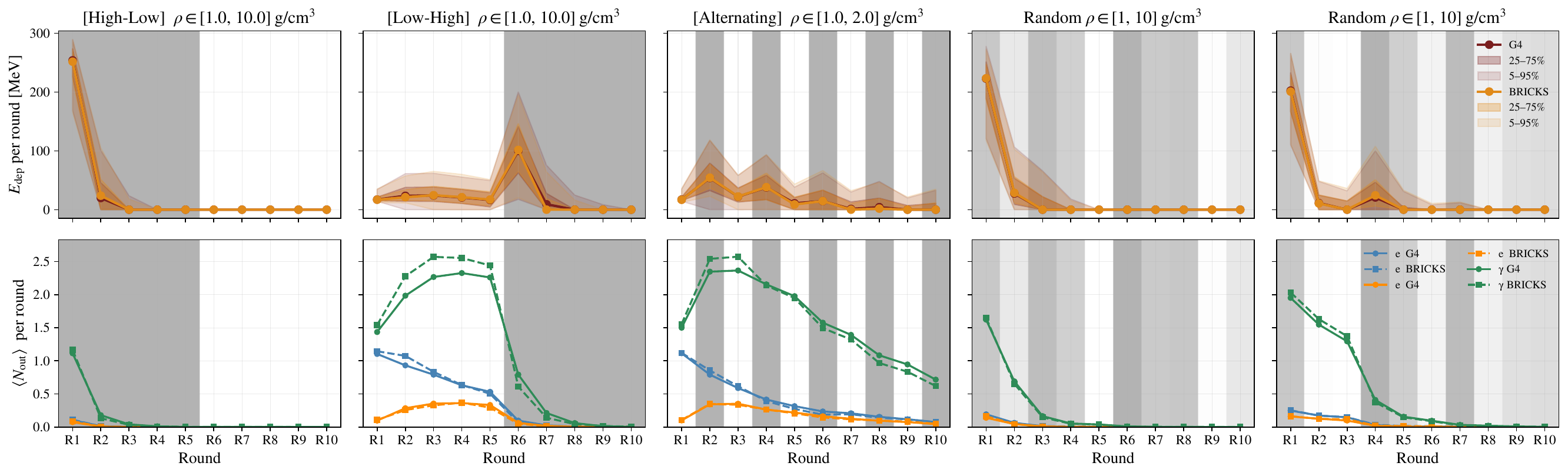}
\vspace{-0.6cm}
\caption{Zero-Shot performance of the energy deposition and particle multiplicity evolution over multi-round autoregressive rollouts for hand-picked (left three) and random (right two) environments. Top: Geant4 (maroon) and BRICKS (orange), bands denote sample variance. Bottom: BRICKS (dashed) and Geant4 (solid).}
\label{fig:composition}
\end{figure}

A core of the simulation model is the composition of multiple kernel invocations at inference time using autoregression. Therefore, single-kernel metrics do not fully capture the quality of the kernel. Instead, the stability of the prediction under multiple rounds of autoregression is crucial. While we leave the development of a fully featured 3D simulation framework based on neural kernels for future work, we can assess performance under composition and thus the model's zero-shot generalization performance in a synthetic setup. The inference time simulation proceeds along the lines described in Section~\ref{sec:probmodel}. In order to compose the model, at each iteration all outgoing particles are parallel-transported to the opposite face to prepare it for the next iteration. I.e. a particle exiting on the $+x$ face (where as an outgoing particle it has a positive inner product with the cube surface normal) is transported to the $-x$ face. The particles now point ``inward'' again with a negative inner product with the surface normal and ready for another iteration through the kernel. The procedure is repeated at each iteration. To demonstrate the zero-shot generalization capabilities with respect to the macroscopic material distribution, we evaluate the composition model in a setting where the density varies from one iteration round to the next. In Figure~\ref{fig:composition} we compare the evolution of the energy depositions and particle multiplicity in five synthetic scenarios, the latter of which are random. The shading corresponds to the material density. High densities induce large energy depositions: if density increases suddenly it induces a large material side effect and the number of particles ``in-flight'' drops rapidly. To quantify the agreement, we concatenate the particle multiplicities and energy depositions of all $N$ rounds into a high-dimensional vector and use it as a summary vector for the MMD metric. In Table~\ref{tab:mmd_composition_summary}, we compare the MMD score of our model against the ground truth and list the statistical floor of the ground truth compared to itself as a reference. In particular ``Random10'' showcases the MMD score observed for \emph{random} material configurations, which are compatible with the hand-picked scenario and indicate stable zero-shot generalization.

  \begin{table}[h]
    \caption{Zero-Shot Performance of iterated BRICKS rollouts as measured by Maximum Mean Discrepancy on rollout summary vectors.}
  \centering
    \label{tab:mmd_composition_summary}
    \vspace{0.1cm}
    \small
    \setlength{\tabcolsep}{6pt}
    \renewcommand{\arraystretch}{1.15}                      
    \begin{tabular}{@{}ll cccc@{}}
      \toprule                                              
      \textsc{Particle} & \textsc{Comparison}
        & High-Low & Low-High & Alternating & Random10 \\
      \midrule                            
      \multirow{2}{*}{Electron}
        & G4--G4 & $0.009 \pm 0.018$ & $0.009 \pm 0.014$ & $0.006 \pm 0.009$ & $0.011 \pm 0.002$ \\
        & FM--G4 & $0.033 \pm 0.019$ & $0.074 \pm 0.007$ & $0.052 \pm 0.006$ & $0.027 \pm 0.014$ \\
      \cmidrule(lr){1-6}
      \multirow{2}{*}{Photon}
        & G4--G4 & $0.014 \pm 0.018$ & $0.008 \pm 0.011$ & $0.012 \pm 0.012$ & $0.009 \pm 0.002$ \\
        & FM--G4 & $0.007 \pm 0.013$ & $0.061 \pm 0.005$ & $0.046 \pm 0.006$ & $0.028 \pm 0.012$ \\
      \bottomrule
    \end{tabular}
  \end{table}

\vspace{-0.4cm}

\vspace{-0.1cm}
\section{Discussion and Outlook}

We formalize a new method for creating surrogate models for radiation-matter interactions. Our model exploits the Markov nature of particle interactions to distill the complex medium-scale physics into an amortized neural representation. Unlike existing surrogate models trained on full rollouts which are tied to a specific matter geometry, our model can be used as a Markov kernel in autoregressive rollouts to generalize zero-shot to unseen geometries composed of a known set of building blocks. 

We introduce a new dataset and implement a concrete reference implementation of our simulation model using a transformer-based mixed discrete-continuous conditional generative model. In evaluation we observe excellent agreement between ground truth distribution from mechanistic simulators and our surrogate model. At a single-kernel level, the per-event GPU wall-clock time shows an acceleration with respect to CPU-bound mechanistic simulators. 

The present work showcases that the overall simulation model is sound but does not yet constitute a full production-grade drop-in replacement for mechanistic simulators. For this a full 3D autoregressive iteration must be developed and the kernel must be extended to cover more complex per-building block environments $\lambda$ and a broader set of incoming and outgoing particles. Our work does, however, provide first insight into the autoregressive stability: our model achieves a sufficiently high precision such that catastrophic error accumulation over multi-step rollouts could be avoided. 

The observed out-of-the-box acceleration of a single kernel invocation compared to an equivalent mechanistic simulation is encouraging considering the choice of flow-matching as an architecture, which, while precise, is not optimized for inference speed. It must be noted that given the relative cost of hardware, further inference-time optimizations at the kernel-level and an eventual outer-loop autoregressive framework will be required in order to realize these gains in production settings. Here, strategies analogous to language models, such as KV caching~\citep{pope2023efficiently} should be explored.


\section*{Acknowledgments}

We gratefully acknowledge early work by and/or discussions with Johann Brehmer, Gilles Louppe, Kyle Cranmer, Annalena Kofler, Nicole Hartman and Sven Menke as well as technical support by Matthew Feickert. The work of R.H. and L.H. was supported by the European Research Council (ERC) under the European Union’s Horizon
Europe research and innovation program grant agreement 101220713 (LEGO). B.H. was by the Excellence Cluster ORIGINS, which is funded by the Deutsche Forschungsgemeinschaft (DFG, German Research Foundation) under Germany’s Excellence
Strategy - EXC-2094-390783311. R.H. was further supported by seed-funding courtesy of the Munich Data Science Institute (MDSI). M.K. is supported by the US Department of Energy (DOE) under Grant No. DE-AC02-76SF00515. Part of this work was  carried out at the Munich Institute for Astro-, Particle and BioPhysics (MIAPbP), which is funded by the Deutsche Forschungsgemeinschaft (DFG, German Research Foundation) under Germany’s Excellence Strategy – EXC-2094 – 390783311. We gratefully acknowledge computing resources provided by the Max-Planck Institute for Physics, the Max-Planck Computing and Data Facility (MPCDF) as well as the Leibniz Rechenzentrum (LRZ).
S.G. is supported by the IIT-Kanpur faculty initiation grant (IITK /PHY /2023499) and Anusandhan National Research Foundation, Advanced Research Grant (ANRF/ARG/2025/005801/PS) from Govt. of India.


\bibliographystyle{unsrtnat}
\bibliography{refs}

@misc{wang2020xtransformers,                              
    author       = {Wang, Phil},
    title        = {x-transformers},                               
    year         = {2020},
    howpublished = {\url{https://github.com/lucidrains/x-transformers}}, 
    note         = {MIT License}      
  }

@techreport{CERN-LHCC-2022-005,
      collaboration = "ATLAS",
      title         = "{ATLAS Software and Computing HL-LHC Roadmap}",
      institution   = "CERN",
      reportNumber  = "CERN-LHCC-2022-005, LHCC-G-182",
      address       = "Geneva",
      year          = "2022",
      url           = "https://cds.cern.ch/record/2802918",
}

@article{pope2023efficiently,
  title={Efficiently scaling transformer inference},
  author={Pope, Reiner and Douglas, Sholto and Chowdhery, Aakanksha and Devlin, Jacob and Bradbury, James and Heek, Jonathan and Xiao, Kefan and Agrawal, Shivani and Dean, Jeff},
  journal={Proceedings of machine learning and systems},
  volume={5},
  pages={606--624},
  year={2023}
}

@article{gretton2012kernel,
  title={A kernel two-sample test},
  author={Gretton, Arthur and Borgwardt, Karsten M and Rasch, Malte J and Sch{\"o}lkopf, Bernhard and Smola, Alexander},
  journal={The journal of machine learning research},
  volume={13},
  number={1},
  pages={723--773},
  year={2012},
  publisher={JMLR. org}
}

@article{szekely2013energy,
  title={Energy statistics: A class of statistics based on distances},
  author={Sz{\'e}kely, G{\'a}bor J and Rizzo, Maria L},
  journal={Journal of statistical planning and inference},
  volume={143},
  number={8},
  pages={1249--1272},
  year={2013},
  publisher={Elsevier}
}

@article{geant4,
    author = "Agostinelli, S. and others",
    collaboration = "GEANT4",
    title = "{GEANT4 - A Simulation Toolkit}",
    reportNumber = "SLAC-PUB-9350, FERMILAB-PUB-03-339, CERN-IT-2002-003",
    doi = "10.1016/S0168-9002(03)01368-8",
    journal = "Nucl. Instrum. Meth. A",
    volume = "506",
    pages = "250--303",
    year = "2003"
}

@article{calochallenge,
    author = "Amram, Oz and others",
    editor = "Krause, Claudius and Faucci Giannelli, Michele and Kasieczka, Gregor and Nachman, Benjamin and Salamani, Dalila and Shih, David and Zaborowska, Anna",
    title = "{CaloChallenge 2022: a community challenge for fast calorimeter simulation}",
    eprint = "2410.21611",
    archivePrefix = "arXiv",
    primaryClass = "physics.ins-det",
    reportNumber = "HEPHY-ML-24-05, FERMILAB-PUB-24-0728-CMS, TTK-24-43",
    doi = "10.1088/1361-6633/ae1304",
    journal = "Rept. Prog. Phys.",
    volume = "88",
    number = "11",
    pages = "116201",
    year = "2025"
}

@misc{defazio2024road,
    title={The Road Less Scheduled}, 
    author={Aaron Defazio and Xingyu Yang and Harsh Mehta and Konstantin Mishchenko and Ahmed Khaled and Ashok Cutkosky},
    year={2024},
    eprint={2405.15682},
    archivePrefix={arXiv},
    primaryClass={cs.LG}
}

@article{KAWTIKWAR2024104838,
    title = {HyLAC: Hybrid linear assignment solver in CUDA},
    journal = {Journal of Parallel and Distributed Computing},
    volume = {187},
    pages = {104838},
    year = {2024},
    issn = {0743-7315},
    doi = {https://doi.org/10.1016/j.jpdc.2024.104838},
    url = {https://www.sciencedirect.com/science/article/pii/S0743731524000029},
    author = {Samiran Kawtikwar and Rakesh Nagi}
}

@misc{luo2026soflowsolutionflowmodels,
    title={SoFlow: Solution Flow Models for One-Step Generative Modeling}, 
    author={Tianze Luo and Haotian Yuan and Zhuang Liu},
    year={2026},
    eprint={2512.15657},
    archivePrefix={arXiv},
    primaryClass={cs.LG},
    url={https://arxiv.org/abs/2512.15657}, 
}

@misc{lipman2024flowmatchingguidecode,
      title={Flow Matching Guide and Code}, 
      author={Yaron Lipman and Marton Havasi and Peter Holderrieth and Neta Shaul and Matt Le and Brian Karrer and Ricky T. Q. Chen and David Lopez-Paz and Heli Ben-Hamu and Itai Gat},
      year={2024},
      eprint={2412.06264},
      archivePrefix={arXiv},
      primaryClass={cs.LG},
      url={https://arxiv.org/abs/2412.06264}, 
}

@article{hashemi2024taxonomic,
    title={Deep Generative Models for Detector Signature Simulation: A Taxonomic Review},
    author={Hashemi, Baran and Krause, Claudius},
    journal={Reviews in Physics},
    volume={12},
    pages={100092},
    year={2024},
    doi={10.1016/j.revip.2024.100092},
    url={https://arxiv.org/abs/2312.09597}
}

@article{paganini2018calogan,
    title={CaloGAN: Simulating 3D High Energy Particle Showers in Multilayer Electromagnetic Calorimeters with Generative Adversarial Networks},
    author={Paganini, Michela and de Oliveira, Luke and Nachman, Benjamin},
    journal={Physical Review D},
    volume={97},
    number={1},
    pages={014021},
    year={2018},
    doi={10.1103/PhysRevD.97.014021},
    url={https://arxiv.org/abs/1712.10321}
}

@misc{krause2021caloflow,
    title={CaloFlow: Fast and Accurate Generation of Calorimeter Showers with Normalizing Flows},
    author={Krause, Claudius and Shih, David},
    year={2021},
    eprint={2106.05285},
    archivePrefix={arXiv},
    primaryClass={hep-ph},
    url={https://arxiv.org/abs/2106.05285}
}

@article{mikuni2024caloscorev2,
    title={CaloScore v2: Single-shot Calorimeter Shower Simulation with Diffusion Models},
    author={Mikuni, Vinicius and Nachman, Benjamin},
    journal={Journal of Instrumentation},
    volume={19},
    number={02},
    pages={P02001},
    year={2024},
    doi={10.1088/1748-0221/19/02/P02001},
    url={https://arxiv.org/abs/2308.03847}
}

@article{favaro2025calodream,
    title={CaloDREAM -- Detector Response Emulation via Attentive Flow Matching},
    author={Favaro, Luigi and Ore, Ayodele and Palacios Schweitzer, Sofia and Plehn, Tilman},
    journal={SciPost Physics},
    volume={18},
    pages={088},
    year={2025},
    doi={10.21468/SciPostPhys.18.3.088},
    url={https://arxiv.org/abs/2405.09629}
}

@article{hashemi2024ultrahigh,
    title={Ultra-high-granularity detector simulation with intra-event aware generative adversarial network and self-supervised relational reasoning},
    author={Hashemi, Baran and Hartmann, Nikolai and Sharifzadeh, Sahand and Kahn, James and Kuhr, Thomas},
    journal={Nature Communications},
    volume={15},
    pages={4916},
    year={2024},
    doi={10.1038/s41467-024-49104-4},
    url={https://www.nature.com/articles/s41467-024-49104-4}
}

@article{diefenbacher2023newangles,
    title={New Angles on Fast Calorimeter Shower Simulation},
    author={Diefenbacher, Sascha and Eren, Engin and Gaede, Frank and Kasieczka, Gregor and Korol, Anatolii and Kr{\"u}ger, Katja and McKeown, Peter and Rustige, Lennart},
    journal={Machine Learning: Science and Technology},
    volume={4},
    number={3},
    pages={035044},
    year={2023},
    doi={10.1088/2632-2153/acefa9},
    url={https://arxiv.org/abs/2303.18150}
}

@article{salamani2023metahep,
    title={MetaHEP: Meta learning for fast shower simulation of high energy physics experiments},
    author={Salamani, Dalila and Zaborowska, Anna and Pokorski, Witold},
    journal={Physics Letters B},
    volume={844},
    pages={138079},
    year={2023},
    doi={10.1016/j.physletb.2023.138079},
    url={https://www.sciencedirect.com/science/article/pii/S0370269323004136}
}

@article{amram2023calodiffusion,
    title={Denoising diffusion models with geometry adaptation for high fidelity calorimeter simulation},
    author={Amram, Oz and Pedro, Kevin},
    journal={Physical Review D},
    volume={108},
    number={7},
    pages={072014},
    year={2023},
    doi={10.1103/PhysRevD.108.072014},
    url={https://arxiv.org/abs/2308.03876}
}

@article{nguyen2026differentiable,
    title={Differentiable Surrogate for Detector Simulation and Design with Diffusion Models},
    author={Nguyen, Xuan Tung and Chen, Long and Dorigo, Tommaso and Gauger, Nicolas R. and Vischia, Pietro and Nardi, Federico and Awais, Muhammad and Hanif, Hamza and Abbas, Shahzaib and Kapoor, Rukshak},
    journal={Machine Learning: Science and Technology},
    volume={7},
    pages={025061},
    year={2026},
    doi={10.1088/2632-2153/ae5c56},
    url={https://arxiv.org/abs/2601.07859}
}

@article{buhmann2023caloclouds,
    title={CaloClouds: Fast Geometry-Independent Highly-Granular Calorimeter Simulation},
    author={Buhmann, Erik and Diefenbacher, Sascha and Eren, Engin and Gaede, Frank and Kasieczka, Gregor and Korol, Anatolii and Korcari, William and Kr{\"u}ger, Katja and McKeown, Peter},
    journal={Journal of Instrumentation},
    volume={18},
    number={11},
    pages={P11025},
    year={2023},
    doi={10.1088/1748-0221/18/11/P11025},
    url={https://arxiv.org/abs/2305.04847}
}

@article{kobylianskii2024calograph,
    title={CaloGraph: Graph-based diffusion model for fast shower generation in calorimeters with irregular geometry},
    author={Kobylianskii, Dmitrii and Soybelman, Nathalie and Dreyer, Etienne and Gross, Eilam},
    journal={Physical Review D},
    volume={110},
    number={7},
    pages={072003},
    year={2024},
    doi={10.1103/PhysRevD.110.072003},
    url={https://arxiv.org/abs/2402.11575}
}

@article{hashemi2024deep,
  title={Deep generative models for ultra-high granularity particle physics detector simulation: A voyage from emulation to extrapolation},
  author={Hashemi, Baran},
  url={https://arxiv.org/abs/2403.13825},
  year={2024}
}

@article{buss2026caloclouds3,
    title={CaloClouds3: Ultra-Fast Geometry-Independent Highly-Granular Calorimeter Simulation},
    author={Buss, Thorsten and Day-Hall, Henry and Gaede, Frank and Kasieczka, Gregor and Kr{\"u}ger, Katja and Korol, Anatolii and Madlener, Thomas and McKeown, Peter and Mozzanica, Martina and Valente, Lorenzo},
    journal={Journal of Instrumentation},
    volume={21},
    number={03},
    pages={P03018},
    year={2026},
    doi={10.1088/1748-0221/21/03/P03018},
    url={https://arxiv.org/abs/2511.01460}
}

@article{erdmann2025paraflow,
    title={Paraflow: fast calorimeter simulations parameterized in upstream material configurations},
    author={Erdmann, Johannes and Kann, Jonas and Mausolf, Florian and Wissmann, Peter and others},
    journal={European Physical Journal C},
    volume={85},
    pages={857},
    year={2025},
    doi={10.1140/epjc/s10052-025-14604-0},
    url={https://link.springer.com/article/10.1140/epjc/s10052-025-14604-0}
}

@inproceedings{sanchezgonzalez2020learning,
    title={Learning to Simulate Complex Physics with Graph Networks},
    author={Sanchez-Gonzalez, Alvaro and Godwin, Jonathan and Pfaff, Tobias and Ying, Rex and Leskovec, Jure and Battaglia, Peter},
    booktitle={Proceedings of the 37th International Conference on Machine Learning},
    series={Proceedings of Machine Learning Research},
    volume={119},
    pages={8459--8468},
    year={2020},
    url={https://proceedings.mlr.press/v119/sanchez-gonzalez20a.html}
}

@inproceedings{brandstetter2022message,
    title={Message Passing Neural PDE Solvers},
    author={Brandstetter, Johannes and Worrall, Daniel E. and Welling, Max},
    booktitle={International Conference on Learning Representations},
    year={2022},
    url={https://openreview.net/forum?id=vSix3HPYKSU}
}

@misc{koehler2024apebench,
    title={APEBench: A Benchmark for Autoregressive Neural Emulators of PDEs},
    author={Koehler, Felix and Niedermayr, Simon and Westermann, R{\"u}diger and Thuerey, Nils},
    year={2024},
    eprint={2411.00180},
    archivePrefix={arXiv},
    primaryClass={cs.LG},
    note={Accepted at NeurIPS 2024},
    url={https://arxiv.org/abs/2411.00180}
}

@misc{farmer2025gmc,
    title={Generative Monte Carlo Sampling for Constant-Cost Particle Transport},
    author={Farmer, Joseph A. and Murray, Aidan and Krotz, Johannes and McClarren, Ryan G.},
    year={2025},
    eprint={2512.13965},
    archivePrefix={arXiv},
    primaryClass={physics.comp-ph},
    url={https://arxiv.org/abs/2512.13965}
}

@misc{xu2019understandingimprovinglayernormalization,
      title={Understanding and Improving Layer Normalization}, 
      author={Jingjing Xu and Xu Sun and Zhiyuan Zhang and Guangxiang Zhao and Junyang Lin},
      year={2019},
      eprint={1911.07013},
      archivePrefix={arXiv},
      primaryClass={cs.LG},
      url={https://arxiv.org/abs/1911.07013}, 
}

@misc{chen2024flowmatchinggeneralgeometries,
      title={Flow Matching on General Geometries}, 
      author={Ricky T. Q. Chen and Yaron Lipman},
      year={2024},
      eprint={2302.03660},
      archivePrefix={arXiv},
      primaryClass={cs.LG},
      url={https://arxiv.org/abs/2302.03660}, 
}

@misc{vaswani2023attentionneed,
      title={Attention Is All You Need}, 
      author={Ashish Vaswani and Noam Shazeer and Niki Parmar and Jakob Uszkoreit and Llion Jones and Aidan N. Gomez and Lukasz Kaiser and Illia Polosukhin},
      year={2023},
      eprint={1706.03762},
      archivePrefix={arXiv},
      primaryClass={cs.CL},
      url={https://arxiv.org/abs/1706.03762}, 
}

@misc{lipman2023flowmatchinggenerativemodeling,
      title={Flow Matching for Generative Modeling}, 
      author={Yaron Lipman and Ricky T. Q. Chen and Heli Ben-Hamu and Maximilian Nickel and Matt Le},
      year={2023},
      eprint={2210.02747},
      archivePrefix={arXiv},
      primaryClass={cs.LG},
      url={https://arxiv.org/abs/2210.02747}, 
}

@inproceedings{SPENVIS,
    author = {Calders, Stijn and Messios, Neophytos and Botek, Edith and De Donder, Erwin and Kruglanski, Michel and Evans, Hugh and Rodgers, David},
    year = {2018},
    month = {05},
    pages = {},
    title = {Modeling the space environment and its effects on spacecraft and astronauts using SPENVIS},
    doi = {10.2514/6.2018-2598}
}

@article{Med_MC,
    author = {Guatelli, S. and Incerti, S.},
    year = {2017},
    month = {01},
    pages = {},
    title = {Monte Carlo simulations for medical physics: From fundamental physics to cancer treatment},
    volume = {33},
    journal = {Physica Medica},
    doi = {10.1016/j.ejmp.2017.01.002}
}

@article{ATLAS:2010arf,
    author = "Aad, G. and others",
    collaboration = "ATLAS",
    title = "{The ATLAS Simulation Infrastructure}",
    eprint = "1005.4568",
    archivePrefix = "arXiv",
    primaryClass = "physics.ins-det",
    doi = "10.1140/epjc/s10052-010-1429-9",
    journal = "Eur. Phys. J. C",
    volume = "70",
    pages = "823--874",
    year = "2010"
}

@article{GARCIA201773,
    title = {New physics model in GEANT4 for the simulation of neutron interactions with organic scintillation detectors},
    journal = {Nuclear Instruments and Methods in Physics Research Section A: Accelerators, Spectrometers, Detectors and Associated Equipment},
    volume = {868},
    pages = {73-81},
    year = {2017},
    issn = {0168-9002},
    doi = {https://doi.org/10.1016/j.nima.2017.06.021},
    url = {https://www.sciencedirect.com/science/article/pii/S0168900217306745},
    author = {A.R. Garcia and E. Mendoza and D. Cano-Ott and R. Nolte and T. Martinez and A. Algora and J.L. Tain and K. Banerjee and C. Bhattacharya}
}

@article{AGOSTINELLI2003250,
    title = {Geant4—a simulation toolkit},
    journal = {Nuclear Instruments and Methods in Physics Research Section A: Accelerators, Spectrometers, Detectors and Associated Equipment},
    volume = {506},
    number = {3},
    pages = {250-303},
    year = {2003},
    issn = {0168-9002},
    doi = {https://doi.org/10.1016/S0168-9002(03)01368-8},
    url = {https://www.sciencedirect.com/science/article/pii/S0168900203013688},
    author = {S. Agostinelli and J. Allison and K. Amako and J. Apostolakis and H. Araujo and P. Arce and M. Asai and D. Axen and S. Banerjee and G. Barrand and F. Behner and L. Bellagamba and J. Boudreau and L. Broglia and A. Brunengo and H. Burkhardt and S. Chauvie and J. Chuma and R. Chytracek and G. Cooperman and G. Cosmo and P. Degtyarenko and A. Dell'Acqua and G. Depaola and D. Dietrich and R. Enami and A. Feliciello and C. Ferguson and H. Fesefeldt and G. Folger and F. Foppiano and A. Forti and S. Garelli and S. Giani and R. Giannitrapani and D. Gibin and J.J. {Gómez Cadenas} and I. González and G. {Gracia Abril} and G. Greeniaus and W. Greiner and V. Grichine and A. Grossheim and S. Guatelli and P. Gumplinger and R. Hamatsu and K. Hashimoto and H. Hasui and A. Heikkinen and A. Howard and V. Ivanchenko and A. Johnson and F.W. Jones and J. Kallenbach and N. Kanaya and M. Kawabata and Y. Kawabata and M. Kawaguti and S. Kelner and P. Kent and A. Kimura and T. Kodama and R. Kokoulin and M. Kossov and H. Kurashige and E. Lamanna and T. Lampén and V. Lara and V. Lefebure and F. Lei and M. Liendl and W. Lockman and F. Longo and S. Magni and M. Maire and E. Medernach and K. Minamimoto and P. {Mora de Freitas} and Y. Morita and K. Murakami and M. Nagamatu and R. Nartallo and P. Nieminen and T. Nishimura and K. Ohtsubo and M. Okamura and S. O'Neale and Y. Oohata and K. Paech and J. Perl and A. Pfeiffer and M.G. Pia and F. Ranjard and A. Rybin and S. Sadilov and E. {Di Salvo} and G. Santin and T. Sasaki and N. Savvas and Y. Sawada and S. Scherer and S. Sei and V. Sirotenko and D. Smith and N. Starkov and H. Stoecker and J. Sulkimo and M. Takahata and S. Tanaka and E. Tcherniaev and E. {Safai Tehrani} and M. Tropeano and P. Truscott and H. Uno and L. Urban and P. Urban and M. Verderi and A. Walkden and W. Wander and H. Weber and J.P. Wellisch and T. Wenaus and D.C. Williams and D. Wright and T. Yamada and H. Yoshida and D. Zschiesche}
}

@article{doi:10.1073/pnas.1915980117,
    author = {Johann Brehmer  and Gilles Louppe  and Juan Pavez  and Kyle Cranmer },
    title = {Mining gold from implicit models to improve likelihood-free inference},
    journal = {Proceedings of the National Academy of Sciences},
    volume = {117},
    number = {10},
    pages = {5242-5249},
    year = {2020},
    doi = {10.1073/pnas.1915980117},
    URL = {https://www.pnas.org/doi/abs/10.1073/pnas.1915980117},
    eprint = {https://www.pnas.org/doi/pdf/10.1073/pnas.1915980117}
}


\newpage
\section*{Appendix}

\subsection*{Ablations}

We test multiple ablations of the model architecture. The CFM model framework allows for two key choices: the base distribution from which the continuous probability flow begins and the choice of coupling during the training.

\subsubsection*{Base Distribution}

A priori, we expected a physics-informed base that is close to the target distribution to aid performance. We therefore designed a physical base (described below) $p_\mathrm{phys}$ that aims to place particles on the opposite face with some user-defined spread controlled by the parameter $\kappa$. We define two choices of $\kappa$: $\kappa=8$ places the particles narrowly across the opposite (raytraced) position of the cube, while $\kappa=1.4$ leads to a near-isotropic distribution. From the evaluations in Table~\ref{tab:prior_fm}, we confirm that the distributional distance $MMD_\mathrm{phys}$ of the physics-informed choice to the target distribution is smaller than the near-isotropic one as indicated by $MMD_\mathrm{iso}$. Surprisingly, after integration, both choices perform similarly with no significant deviation. This speaks to the expressiveness of the CFM model. We still believe such a physics-informed choice may be useful in future iterations.

\subsubsection*{Naive vs OT Coupling}

We observe a similar result in the choice of coupling. In conditional flow matching, the freedom to choose coupling can help to significantly reduce the variance of the conditional velocity which serves as supervision signal to the neural velocity field prediction, e.g. by pairing similar base and target samples with an optimal transport (OT) map. Here again, surprisingly, this did not significantly improve results. We still keep OT coupling in the model as we expect it to be relevant in the future, but note that this will require more investigations.

\subsection*{Physical Base}
\begin{itemize}

\item Energy Deposition $E_\mathrm{dep} =  \frac{E_\mathrm{in}}{\sum_{n=1}^N E_\mathrm{out}} \in \mathbb{R}: \quad \sim \sigma\left(\mathcal{N}(0, 1)\right)$

\item Momentum magnitude as fraction of the incoming magnitude $\frac{||\vec{p}||_2}{||\vec{p}_\mathrm{in}||_2} \; (\approx \frac{E}{E_\mathrm{in}}) \in \mathbb{R}$: \\
$$ \sim E_\mathrm{upper} \cdot \left(1 - \tanh\left(\frac{1}{2} \cdot |\mathcal{N}(0, 1)|\right)\right) + 1 \qquad \mathrm{with} \; E_\mathrm{upper} = \ln \left(\frac{||\vec{p}_\mathrm{in}||_2}{E_\mathrm{cutoff}}\right)$$

\item Direction $\hat{p}, \hat{x} \in \mathcal{S}^2$: \par
1.: Ray-trace the incoming particle through the cube: \\ $$\vec{p}_\mathrm{ray} = \vec{p}_\mathrm{in}, \qquad \vec{x}_\mathrm{ray, \, cube} = \vec{x}_\mathrm{in, \, cube} + \vec{p}_\mathrm{in} \cdot \min \left((\mathrm{sgn}(\vec{p}_\mathrm{in}) - \vec{x}_\mathrm{in, \, cube}) \oslash {\vec{p}_\mathrm{in}} \right)$$ \\
($\vec{x}_\mathrm{\dots, \, cube}$ is assumed to lie on the unit cube, all computations except $\min$ are element-wise.) \par 

2.: Sample base at $\theta = 0$ pole on $\mathcal{S}^2$: \\
$$\theta_\mathrm{base, \, pole} \sim \pi \cdot \left|\tanh \left( \frac{1}{\kappa} \mathcal{N}(0, 1)\right)\right| \qquad \phi_\mathrm{base, \, pole} \sim \mathcal{U}(0, 2\pi)$$ \\
with a scaling factor $\kappa$. \par

3.: Rotate base samples $(\theta_\mathrm{base, \, pole}, \; \phi_\mathrm{base, \, pole})$ by $(\theta_\mathrm{ray}, \; \phi_\mathrm{ray})$ using the spherical law of cosines.
\end{itemize}

\subsection*{Multiplicity Model Specifics}

The multiplicity model uses a \texttt{ContinuousTransformerWrapper} from the \texttt{x-transformers} package \citep{wang2020xtransformers} to wrap around a \texttt{Decoder} backbone block. Enabled features are \texttt{ff\_swish}, \texttt{ff\_glu}, \texttt{attn\_qk\_norm}, \texttt{rotary\_xpos}, \texttt{use\_adaptive\_rmsnorm}, \texttt{use\_adaptive\_layerscale}, \texttt{residual\_attn}. The used models have a hidden dimension of 128, 6 layers, 6 attention heads, a 4-times feed-forward multiplier, an overall dropout of 0.1 and have a sequence length of 15 (14 particle species). The model was trained on a batch-size of $2^{12}$ over 100 epochs with 20M total events (simulations) and incoming particle types $\in [e^-, e^+, \gamma]$ (thus other particles are filtered from the output when composing multiple kernels). We chose to train with the \texttt{AdamWScheduleFree} optimizer from \texttt{schedulefree} \citep{defazio2024road} with a learning rate of $10^{-3}$, a weight decay of $10^{-2}$ and $\beta = (0.95, 0.999)$. 

\subsection*{Flow Matching Model Specifics}    

The flow matching model uses a \texttt{ContinuousTransformerWrapper} from the \texttt{x-transformers} package \citep{wang2020xtransformers}  to wrap around an \texttt{Encoder} backbone block. The enabled features are \texttt{ff\_swish}, \texttt{ff\_glu}, \texttt{attn\_qk\_norm}, \texttt{attn\_value\_rmsnorm}, \texttt{use\_adaptive\_rmsnorm}, \texttt{use\_adaptive\_layerscale}, \texttt{residual\_attn}. The Used models have a hidden dimension of $2^{9}$ ($2^{8}$ for the smaller model), 6 layers, 8 attention heads, a 4-times feed-forward multiplier and an overall dropout of $0.02$. The flow is defined on the product manifold $\mathbb{R}^3 \times \mathcal{S}^{2}$ (using manifold projections from the \texttt{flow\_matching} package \citep{lipman2024flowmatchingguidecode}),with a base concentration of $\kappa = 8$, a HyLAC \citep{KAWTIKWAR2024104838} style batched OT-coupling solver, and solution consistency loss scale \citep{luo2026soflowsolutionflowmodels} of $1 \cdot 10^{-3}$. The model was trained on a batch-size of $2^{12}$ over 100 epochs with 20M total events (simulations) and incoming particle types $\in [e^-, e^+, \gamma]$. We chose to train with the \texttt{AdamWScheduleFree} optimizer from \texttt{schedulefree} \citep{defazio2024road} with a learning rate of $5 \cdot 10^{-4}$, a weight decay of $10^{-2}$ and betas = (0.95, 0.999).

\subsection*{Evaluation Specifics}

The summary statistic we use to compute the sample-based distance measures is computed as follows: The fixed-size summary statistic $\mathbf{s} \in \mathbb{R}^{34}$ is constructed per event as follows. For each outgoing-particle species $t \in {e^-, e^+, \gamma}$ separately, we compute the per-event mean and standard deviation across the active outgoing-particle slots of five kinematic features: the momentum polar angle $\theta_p$, the momentum azimuthal angle $\phi_p$, the momentum magnitude $|\vec{p}|$, the position polar angle $\theta_x$, and the position azimuthal angle $\phi_x$ (positions are converted to spherical coordinates after the $\mathcal{S}^2$-cube identification described in Section~\ref{sec:probmodel}). This yields $3 \times (5 + 5) = 30$ per-species kinematic moments. We additionally include four event-level scalars: the deposited energy $E_\text{dep}$ and the per-species cardinalities $n_{e^-}, n_{e^+}, n_\gamma$, giving 34 features in total. Per-species statistics are computed only over particles with the corresponding PDG identifier; species absent from a given event contribute zero counts, and the per-species means/standard deviations default to zero in that case. This summary vector serves as input to the MMD and Energy Distance metrics reported in Table~\ref{tab:prior_fm} and Table~\ref{tab:mmd_composition_summary}.

\subsection*{Likelihood Evaluation}

\begin{figure}
    \centering
    \includegraphics[width=0.7\linewidth]{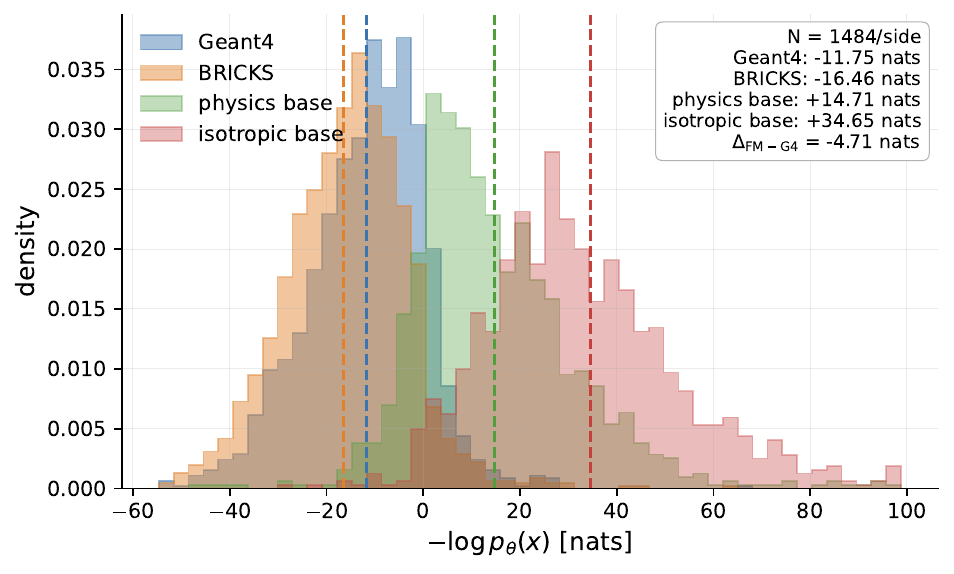}
    \caption{NLL of samples from BRICKS and Geant4 under the BRICKS likelihood}
    \label{fig:lhood}
\end{figure}

While we leave a detailed assessment of the benefit of tractable likelihoods for future work, we show below that the model indeed provides access to the likelihood and behaves as expected. In Figure~\ref{fig:lhood}, the CFM likelihood are shown for samples from the model itself (in orange) and from the mechanistic simulator Geant4. The BRICKS samples evaluated under the BRICKS likelihood achieve a slightly lower negative log-likelihood, but the two distributions are similar indicating a small distributional distance. In future iterations we will explore the use of NLL as a metric as well as use it for downstream tasks. In contrast, the base distributions show a much larger gap to both Geant4 as well as the integrated BRICKS model, further contextualizing the performance.

\subsection*{Timing Methodology}

We measure the speed--quality Pareto curve of \textsc{Bricks-M} (32.5\,M flow
parameters) and \textsc{Bricks-S} (9.76\,M flow parameters) against Geant4 on a
$3\times 5$ grid of Riemannian flow-matching solver settings: three integrators
(Euler, midpoint, RK4) at five step sizes
$h\in\{0.5, 0.25, 0.1, 0.05, 0.025\}$. Each $(\text{method},h)$ cell is reported
as a single point on the $(\bar{t},\mathrm{MMD})$ plane.

For target sample size $N{=}10{,}000$ and batch $B$, we issue
$\lceil N/B\rceil$ batched simulator calls bracketed by
\texttt{torch.cuda.synchronize()} and \texttt{time.perf\_counter()}, and report
per-event time $\bar{t}=\bigl(\sum_i \Delta t_i\bigr)/N$. One warm-up call per
cell is discarded to absorb CUDA kernel/JIT initialization. We repeat each cell
$K{=}5$ times independently and report mean $\pm$ sample standard deviation
(Bessel-corrected, $\mathrm{ddof}{=}1$) along both axes; error bars on the
Pareto plots are these $K$-fold spreads.

\subsection*{Hardware}

All experiments were run on a server equipped with two
Intel\textsuperscript{\textregistered} Xeon\textsuperscript{\textregistered}
Platinum 8568Y+ CPUs (48 cores / 96 threads per socket, 192 threads total)
and an NVIDIA H200 GPU (141\,GB HBM3e, Hopper architecture),
running CUDA 13.0 with driver 580.126.09. Memory usage was approx. 40\,GB for the larger model and approx. 20\,GB for the smaller model, the 100 epochs completed in approx. 20\,h and 10\,h respectively. 

\end{document}